\documentclass[runningheads]{llncs}
\usepackage{graphicx}
\usepackage{comment}
\usepackage{amsmath,amssymb} 
\usepackage{color}

\usepackage{times}
\usepackage{epsfig}
\usepackage{graphicx}
\usepackage{amsmath}
\usepackage{amssymb}
\usepackage{booktabs}   

\usepackage{adjustbox}
\usepackage{color}
\usepackage{graphicx}
\usepackage{xspace}
\usepackage{amssymb}
\usepackage{multirow}
\usepackage{multicol}
\usepackage{algorithm}
\usepackage{algpseudocode}
\usepackage{colortbl}
\usepackage{float}
\usepackage{subfigure}

\newcommand{\model}{TFG\xspace}

\newcommand\minisection[1]{\vspace{1mm}\noindent \textbf{#1}}

\begin{document}
\pagestyle{headings}
\mainmatter
\def\ECCVSubNumber{6913}  

\title{Task-Aware Feature Generation for Zero-Shot Compositional Learning} 

\titlerunning{Task-Aware Feature Generation for Zero-Shot Compositional Learning}
%
\author{Xin Wang \and Fisher Yu \and Trevor Darrell \and Joseph E. Gonzalez}
\authorrunning{Wang et al.}
%
\institute{UC Berkeley}
\maketitle

\begin{abstract}
Visual concepts (e.g., red apple, big elephant) are often semantically compositional and
each element of the compositions can be reused to construct novel concepts (e.g., red elephant). Compositional feature synthesis, which generates image feature distributions exploiting the semantic compositionality, is a promising approach to sample-efficient model generalization. In this work, we propose a task-aware feature generation (TFG) framework for compositional learning, which generates features of novel visual concepts by transferring knowledge
from previously seen concepts. 
These synthetic features are then used to train a classifier to recognize novel concepts
in a zero-shot manner. Our novel \model design injects task-conditioned noise layer-by-layer, producing task-relevant variation at each level. We find the proposed generator design improves
classification accuracy and sample efficiency. Our model
establishes a new state of the art on three zero-shot compositional learning (ZSCL) benchmarks, outperforming the previous discriminative models by a large margin. Our model improves the performance of the prior arts by over 2$\times$ in the generalized ZSCL setting. 
\end{abstract}

\section{Introduction}
Recognizing a vast number of visual concepts, which often follow a long-tail distribution~\cite{salakhutdinov2011learning,wang2017learning,liu2019large}, is a daunting challenge for machine visual systems. Conventional approaches typically require massive training data~\cite{he2017mask,he2016deep,girshick2015fast} which is costly and sometimes even impossible to annotate. However, humans have little problem understanding a rare concept such as \textit{red elephant}, even though they
may not have observed a real red elephant before. They can imagine what a {red elephant} might
look like based on past knowledge about \textit{red} objects and \textit{elephants}. Endowing machines with a similar
ability to synthesize images or even features of novel visual concepts could alleviate the data scarcity issue
and enhance model generalization in low data regimes~\cite{felix2018multi,gao2018low,hariharan2017low,wang2018low,xian2018feature}.

\begin{figure}
    \centering
    \includegraphics[width=.6\linewidth]{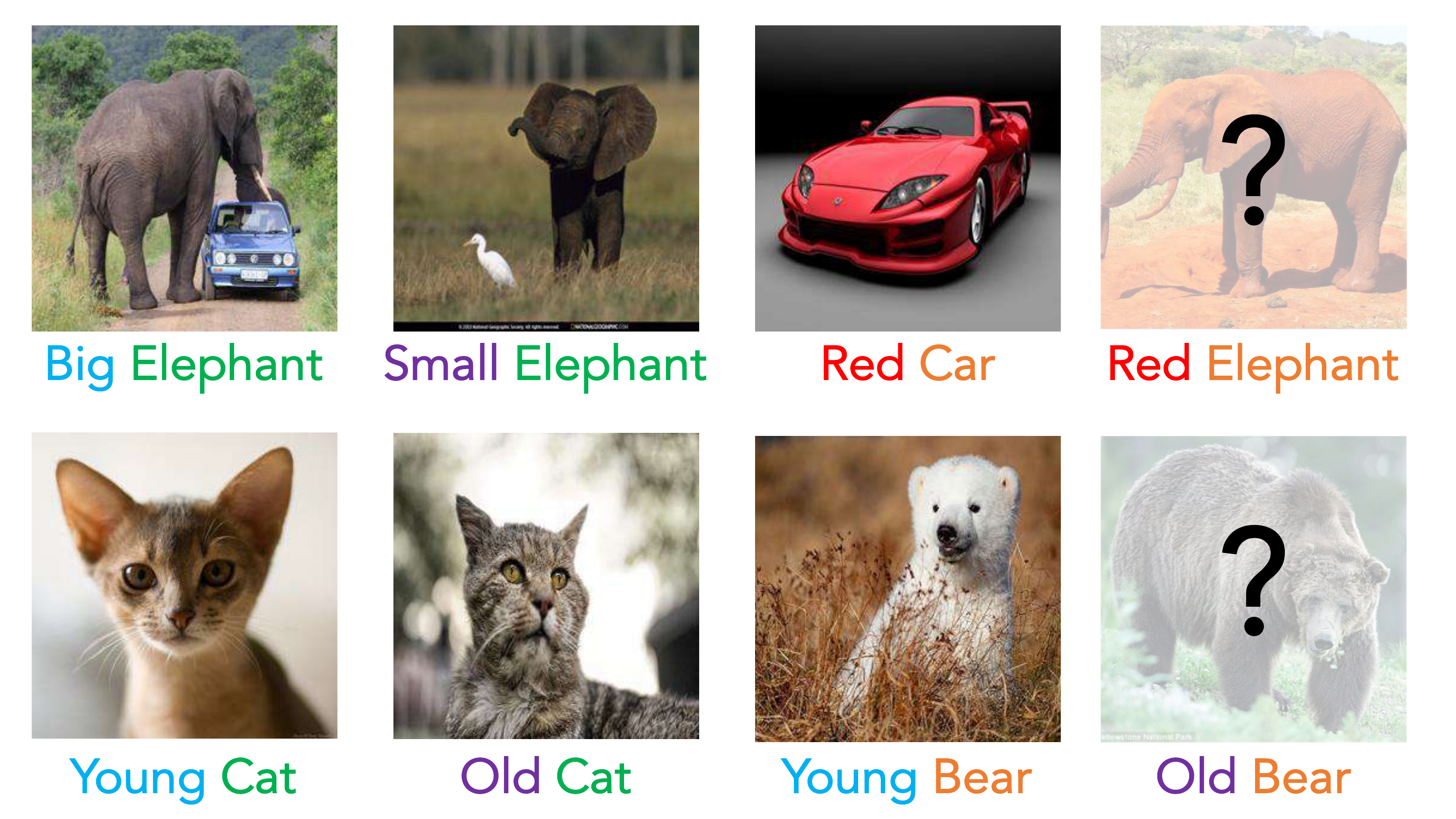}
    \caption{The task of zero-shot compositional learning is to build a classifier for recognizing visual concepts represented by an attribute-object pair (e.g., \texttt{old bear}) where no training images of the composition are available. Our model generates synthetic features for novel compositions, transferring knowledge from the observed compositions (e.g., \texttt{old cat}, \texttt{young bear}). The synthetic features are used for training the classifier directly. \vspace{-5mm}
    \label{fig:fig1}}
\end{figure}

 One way to handle novel visual concepts is to exploit their {compositionality}~\cite{misra2017red}. The machine may not have observed any example of \textit{red elephant} (novel composition) while there might be many images of
\textit{red apple} or \textit{red tomato}, as well as \textit{big elephant} (seen compositions) available during training. In this paper, we study the zero-shot compositional learning task, where the model needs to recognize novel attribute-object compositions of which no training images are available by transferring knowledge from seen compositions. 

Constructing task-conditional feature representations has been largely adopted in the recent 
work~\cite{misra2017red,wang2019tafe,nagarajan2018attributes,purushwalkam2019task}. 
Wang \textit{et al.}~\cite{wang2019tafe} and Purushwalkam \textit{et al.}~\cite{purushwalkam2019task} adopt
task-aware modular feature representation by re-configuring the network conditioned on the
attribute-object compositions. Nagarajan
and Grauman~\cite{nagarajan2018attributes} propose to use attributes as operators to modify 
the object features. However, these approaches rely on the compatibility between multi-modal
inputs (i.e. image features and task descriptions in the form of word embeddings), ignoring
the transferable modes of variation (e.g., lightning changes, translation).

In contrast to the existing works, we take a generative perspective, focusing on
feature synthesis for compositional learning. The key hypothesis is that if a generator is capable of synthesizing feature distributions of
the seen compositions, it may transfer to novel compositions, producing synthetic features that are informative enough to train a classifier without needing real images. 

\begin{figure*}[t]
    \centering
    \includegraphics[width=\linewidth]{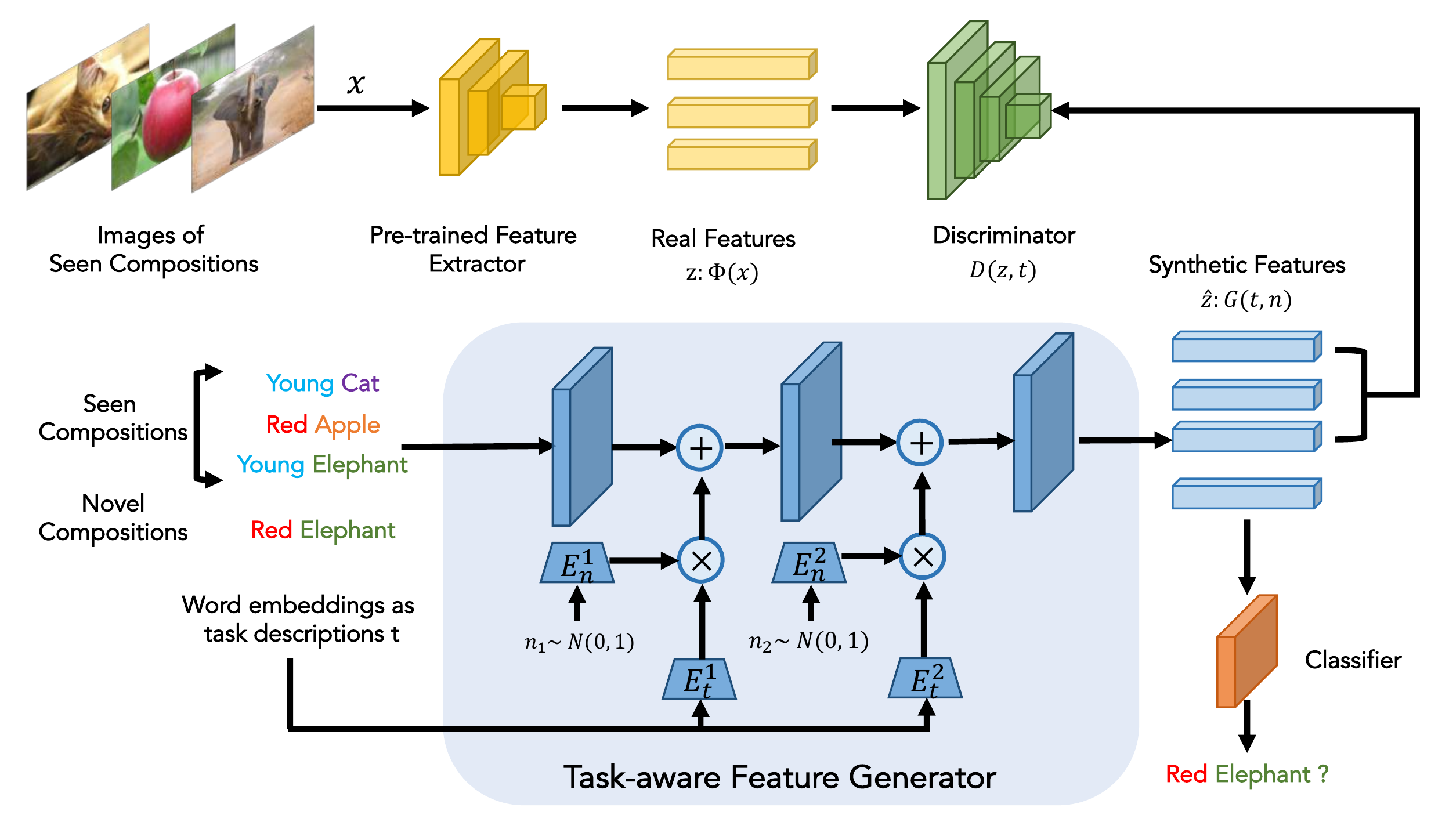}
    \caption{Our approach is composed of a task-aware feature generator (\model), a discriminator and a classifier. 
    During training, \model synthesizes features conditioned on the word 
    embeddings of the composition, which is used to train a classifier for 
    recognizing the seen and novel compositions. The discriminator is introduced
    to distinguish the real and synthetic features of the seen compositions. At inference, we directly use the trained classifier for classification as if it
    were trained with real features. }
    \label{fig:arch}
\end{figure*}

To this end, we propose a task-aware feature generation approach, consisting of a task-aware feature 
generator (\model), a discriminator, and a classifier (Figure~\ref{fig:arch}). During training, the feature
generator synthesizes features conditioned on the word embeddings of the compositions (namely, task
descriptions) and the discriminator is trained to distinguish the synthetic and real features of the seen
compositions. The classifier is jointly trained to recognize the novel compositions using only the synthetic
features. During inference, we just use the trained classifier to directly
recognize features of novel compositions as if it were trained on the real features. 

We introduce \emph{task-aware deep sampling} in our \model model, where the generator adopts the task
description as input and task conditional randomness is incorporated incrementally at each
level. This captures the target image feature distribution efficiently and generates
synthetic features improving generalization of the downstream classifier. Intuitively, samples from task-conditional
distributions are injected layer-by-layer, producing task-relevant variation at each level to
improve sample efficiency and reduce the risk of mode collapse compared to commonly used
generator designs; the latter employ \emph{shallow sampling} where noise is sampled only once at the input
layer of the generator. 

We extensively evaluate our method on three benchmark datasets: MIT-States, UT-Zap50K and StanfordVRD. It outperforms the previous
methods by a large margin. On the new data splits
introduced by the recent paper~\cite{purushwalkam2019task}, our method is able to improve the
performance of the prior art by over 2$\times$.  Moreover, we conduct various architectural
ablation studies as well as qualitative analysis to show that \model is able to effectively
capture the image feature distribution and improve the classification accuracy when
generalizing to novel concepts.


\section{Related Work}
\minisection{Compositional learning.} The idea of compositionality in computer vision can be traced back to the Parts of Recognition work by Hoffman and Richards~\cite{hoffman1984parts}. 
In traditional computer vision, models with pictorial structures~\cite{felzenszwalb2009object,zhu2007stochastic,ikizler2008searching} have been widely studied. 
The compositional learning task, which composes visual primitives and concepts, has been brought back to the deep learning community recently~\cite{tulsiani2017learning,misra2017red,kato2018compositional,andreas2016neural,johnson2017inferring}.  

Misra \textit{et al.} ~\cite{misra2017red} standardized the zero-shot compositional learning task, focusing on classification of images with novel concept compositions during inference.  Several
recent methods~\cite{wang2019tafe,nagarajan2018attributes,purushwalkam2019task} have
exploited the compositional nature of the task by constructing task-aware feature
embeddings. When generalizing to novel concepts, these methods rely on the scoring 
mechanism~\cite{lecun2006tutorial} to leverage the compatibility of the task description in the form of word embeddings and image features. We tackle the problem with task-aware feature generation, synthesizing the image feature distribution of the novel concepts exploiting the semantic compositionality  to transferthe knowledge learned from the seen compositions. 

\minisection{Feature generation.} Prior approaches to data hallucination~\cite{wang2018low,hariharan2017low,gao2018low} and feature 
generation~\cite{xian2018feature,chen2018zero,felix2018multi} have explored the use of the synthetic
features to improve model generalization in the classic zero-/few-shot learning. To the best of our knowledge, we are the first to study feature generation for zero-shot computational learning, which largely exploits the semantic compositionality of visual concepts. In the literature, 
several~\cite{xian2018feature,gao2018low}
have adopted generative adversarial networks (GANs) for feature generation. The closest to our model is 
the feature generation network CLSWGAN proposed by Xian \textit{et al.}~\cite{xian2018feature}.
Our work differs from existing methods in that it explicitly uses the compositional nature of the tasks to interpret the image feature space and introduces a more sample efficient generator design for compositional feature synthesis.
The generator design in CLSWGAN is based on the shallow sampling strategy, which differs from the task-aware deep sampling scheme used in our model. To evaluate the effectiveness of our generator design, we replace the generator used in CLSWGAN~\cite{xian2018feature} with 
ours, achieving faster convergence rate on the zero-shot benchmarks used in~\cite{xian2018feature}.

\minisection{Image generation.} A parallel line of
research~\cite{reed2016generative,han2017stackgan,xian2018feature,yan2016attribute2image} has been
studying conditional image generation using GANs. These papers focus on generating photo-realistic images while in our work, the main goal is to synthesize
informative features instead of images to assist the classifier for recognition tasks. As discussed by Luc \textit{et al.}~\cite{luc2017predicting}, modeling raw RGB intensity overly complicates the task compared to synthesizing high-level scene properties through image features. The latter is not only sufficient but better than 
predicting by directly using raw images for many applications (e.g., semantic segmentation). Our work follows Luc \textit{et al.}~\cite{luc2017predicting} and synthesizes high-level features.

\minisection{Generator designs.} GAN generator designs used for feature synthesis lag behind present research on GANs since previous works~\cite{gulrajani2017improved,miyato2018cgans,brock2018large,lucic2018gans,mescheder2018training} emphasize GAN loss design,
regularization and hyper-parameters to stabilize GAN training. The commonly used generator for
feature synthesis adopts \emph{shallow sampling}, which suffers severely from mode
collapse~\cite{salimans2016improved} and requires substantial training samples in order to
capture the target distribution despite various tricks. StyleGAN~\cite{karras2018style} 
emphasizes the importance of the generator design, proposing a deep
sampling scheme that injects style codes together with randomly sampled noise unconditional
to the task to the generator at each level. StyleGAN improves the quality of the generated 
images due to the advanced generator design. We take a step forward by injecting task-conditional randomness at each level, which we find leads to alleviated mode collapse 
and improved sample efficiency for feature synthesis.

\section{Zero-Shot Compositional Feature Synthesis}
\label{sec:model}

The concept of \emph{compositionality}, which can be traced back to the early
work by Hoffman and Richards~\cite{hoffman1984parts}, is fundamental to 
visual recognition and reasoning. In zero-shot compositional learning,  a key
goal is to exploit compositionality in feature 
learning by transferring knowledge from the seen compositions to novel compositions. 

More formally, we are given a vocabulary of attributes $a\in\mathcal{A}$ and
objects $o\in\mathcal{O}$ as well as a set of image features $\Phi(\mathcal{X})$ extracted by some pre-trained feature extractors (e.g., ResNet~\cite{he2016deep}). A visual concept (a.k.a. category) is represented as an attribute-object pair $c=(a, o) \in \mathcal{C}$ and each image is associated with one composition $c$. 
Moreover, $\mathcal{C}=\mathcal{S}\cup\mathcal{U}$, where the images in the training set associate with the compositions in $\mathcal{S}$ 
and not with the compositions in $\mathcal{U}$. We refer compositions in $S$ as the seen compositions and the 
compositions in $\mathcal{U}$ as novel compositions. Following the tradition
of classic zero-shot learning~\cite{xian2017zero}, the goal is to build a
classifier $f$ which classifies an image feature 
$z:\Phi(x)\in \mathcal{Z}$ using the labels in set of the novel compositions $c\in\mathcal{U}$ (close world setting) 
or using the labels in the set of all compositions $c\in\mathcal{C}$ (open world setting). We use the
concatenation of word embeddings ($t\in\mathcal{T}$) 
of each attribute-object pair $c$ as the task description for 
recognizing the composition $c$ and $\mathcal{T}$ is available during
training.

\subsection{Task-Aware Feature Generation}
We view the zero-shot compositional learning task from the generative 
modeling perspective. The key insight is to learn a projection from
the semantic space  $\mathcal{T}$ to the image
feature space $\mathcal{Z}$ via feature synthesis, rather than projecting the two 
sources of inputs ($z$ and $t$) 
independently into one common embedding space and 
building a model to leverage the compatibility between the two
modalities~\cite{wang2019tafe,nagarajan2018attributes,misra2017red}. 

We now introduce our task-aware feature generator design $G:\mathcal{T}\rightarrow\mathcal{Z}$ for image feature synthesis. 
As illustrated in Figure~\ref{fig:arch}, the task description
$t\in\mathbb{R}^d$ ($d=600$ using the GloVe~\cite{pennington2014glove} to
obtain the word embedding of the compositon) is used as the input to $G$ (instantiated as a stack of fully-connected (FC) layers). At the $i$-th layer
of $G$, random Gaussian noise $n_i\sim\mathcal{N}(0, 1)$ is sampled and then transformed by a sub-network of 2 FC layers, $E_n^i$, obtaining transformed noise $E_n^i(n_i)$. The task description $t$ transformed by $E_t^i$ (also a single FC layer) is multiplied 
with the transformed $E_n^i(n_i)$ to obtain the task-conditioned noise, which is then added to the immediate output of the $i$-th layer of $G$.
Specifically, $\hat{z}^{i+1}$, the input of the $(i+1)$-th layer of $G$, is 
obtained by 
\begin{equation}
    \hat{z}^{i+1} = \hat{z}^i + E_t^i(t) * E_n^i(n_i).
\end{equation}

The feature synthesis procedure can be viewed as using \emph{task-aware deep sampling};
different sets of task-conditioned noise are sampled at different levels of the generator, which
progressively injects task-driven variation to the immediate features of the generator. Intuitively,
the noise injected to the generator is sampled from a task constrained space, which reduces the 
number of samples necessary to learn the projection from the task space $\mathcal{T}$ to the image feature 
space $\mathcal{Z}$. We empirically show that our generator design has better sample efficiency
than the alternative in the experiment section. 

\subsection{Overall Objective}
\label{sec:objective}
Our overall model pipeline is composed of a generator $G$, a discriminator $D$ and a classifier $f$ as illustrated in Figure~\ref{fig:arch}. 
The discriminator is used during training to distinguish whether the input feature of seen compositions is real or fake. We use a simple logistic regression model as our
classifier. All three components are jointly trained in an end-to-end manner and only
the trained classifier is used during testing. The overall objective is 
described below. 

\minisection{Classification loss.} The synthetic features are tailored to help the 
classifier generalization. We include a typical multi-class cross-entropy loss part 
of the objective function. Specially, the classifier $f$ takes the synthetic features
$\hat{z}=G(t, n)$ as input ($n = \{n_i\sim\mathcal{N}(0, 1)| i=1\dots K\}$, $K+1$ is the number of layers of the generator) and output the class prediction $\hat{y} = f(\hat{z})$. The classification loss is defined as 
\begin{equation}
    \mathcal{L}_\text{cls} = -\mathbb{E}_{\hat{z}\sim p_{\hat{z}}}[\log P(y|\hat{z};\theta)],
\end{equation}
where $y$ is ground truth composition that associated with the task description $t$. 
$P(y|\hat{z};\theta)$ is the conditional probability predicted by the classifier $f$
parameterized by $\theta$. 

\minisection{Adversarial training.}  We include a GAN loss to help train the 
generator. We extend the WGAN~\cite{gulrajani2017improved} by integrating the task
descriptions $t$ to both the generator and the discriminator. The extended WGAN loss
can be defined as 
\begin{equation}
    \mathcal{L}_\text{wgan} = \mathbb{E}_{z\sim p_r}[D(z, t)] - \mathbb{E}_{\hat{z}\sim p_g}[D(G(t, n), t)],
\end{equation}
which approximates the Wasserstein distance commonly used in the GAN literature to 
improve training stability compared to the original GAN
loss~\cite{goodfellow2014generative}.  $p_r$ and $p_g$ denote the real feature
distribution and the generated feature distribution respectively. 
We add gradient penalties to the discriminator
to enforce the discriminator to be a 1-Lipschitz function following~\cite{gulrajani2017improved,xian2018feature}. The overall adversarial loss
is defined as 
\begin{equation}
    \mathcal{L}_\text{adv} = \mathcal{L}_\text{wgan} - \lambda_\text{gp}\mathbb{E}(\|\bigtriangledown_{\Tilde{z}}D(\Tilde{z}, t)\|_2 - 1)^2],
\end{equation}
where  $\Tilde{z} = \alpha z + (1-\alpha)\hat{z}$ with $\alpha \sim \text{Uniform}(0,1)$. Following~\cite{gulrajani2017improved,xian2018feature}, we set
$\lambda_\text{gp}=10$ in our experiments. 

Under the zero-shot learning context, only image features of the seen compositions $S$
are available during training; therefore, the adversarial loss $\mathcal{L}_\text{adv}$ is only applied to the seen compositions. 

\minisection{Clustering loss.} To circumvent the challenges of estimating the image feature
distribution, we add a regularization term to make the synthetic features of the seen
compositions closer to the cluster center of the true feature distribution. Intuitively, 
in the extreme case where no randomness is introduced to the generator, $G$ learns a
mapping from $t$ to a ``prototypical'' image feature $\bar{z}\in\mathcal{Z}$. We find this 
regularization term reduces the complexity of modeling the target image feature distribution.

We realize the mapping by introducing a soft-clustering term with $L_2$ regression loss. Specifically, we randomly 
sample a real image feature $z$ of the composition $c\in\mathcal{S}$, and regularize the 
generated feature $\hat{z}$ to be close to $z$. By sampling multiple image features, we 
regularize the generated feature closer to the cluster center of the real image feature distribution. Similar to $\mathcal{L}_\text{adv}$,  the prototypical loss term, defined as 
\begin{equation}
\mathcal{L}_\text{cluster} = \sum_k^K\|\hat{z}_k - z_k\|^2,
\end{equation}
where $K$ features from the seen composition $c$ are sampled. 

\minisection{Overall objective.} 
The overall objective is a weighted sum of the three components shown as 
\begin{equation}
    \min_{G,C}\max_D \mathcal{L}_\text{wgan} + \lambda\mathcal{L}_\text{cls} + \mu\mathcal{L}_\text{cluster},
\end{equation}
and we adopt $\lambda=0.01$ and $\mu=10$ if not specified. Ablations of $\lambda$ and $\mu$ are provided in the supplementary material. 

\subsection{Training and Testing}
Figure~\ref{fig:arch} shows the different components of the our model: the generator $G$, the discriminator $D$ and the classifier $f$. The generator $G$ and the discriminator $D$ are only
used to assist the training of the downstream classifier. If $G$ can generate samples that 
capture the data distribution of the novel composition by transferring the knowledge from the 
seen compositions, a classifier trained with the synthetic features should generalize to the 
real features of the novel compositions during testing. To this end, we train all three
components ($G$, $D$ and $f$) jointly and during testing, we directly feed the real features
of the novel compositions extracted by the pretrained feature extractor to the trained 
classifier.

\section{Experiments}
\label{sec:exp}
We present the experimental evaluation of \model on three zero-shot compositional
learning (ZSCL) benchmark in Section~\ref{sec:composition}. Our method outperforms the previous discriminative models by a large margin. In ection~\ref{sec:gcz}, we evaluate
our model on the new data splits introduced by the recent work~\cite{purushwalkam2019task} in the generalized ZSCL setting. We find our model is able to improve the previous methods by over 2$\times$, establishing a new state of the art. We provide various ablation studies of our proposed generator architecture and loss objective (Section~\ref{sec:ablation}) and a qualitative analysis on the task-aware noise and synthetic features to support understanding our method (Section~\ref{sec:vis}). 

\subsection{Zero-Shot Compositional Learning}
\label{sec:composition}

\begin{figure*}[t]
\centering
\includegraphics[width=\linewidth]{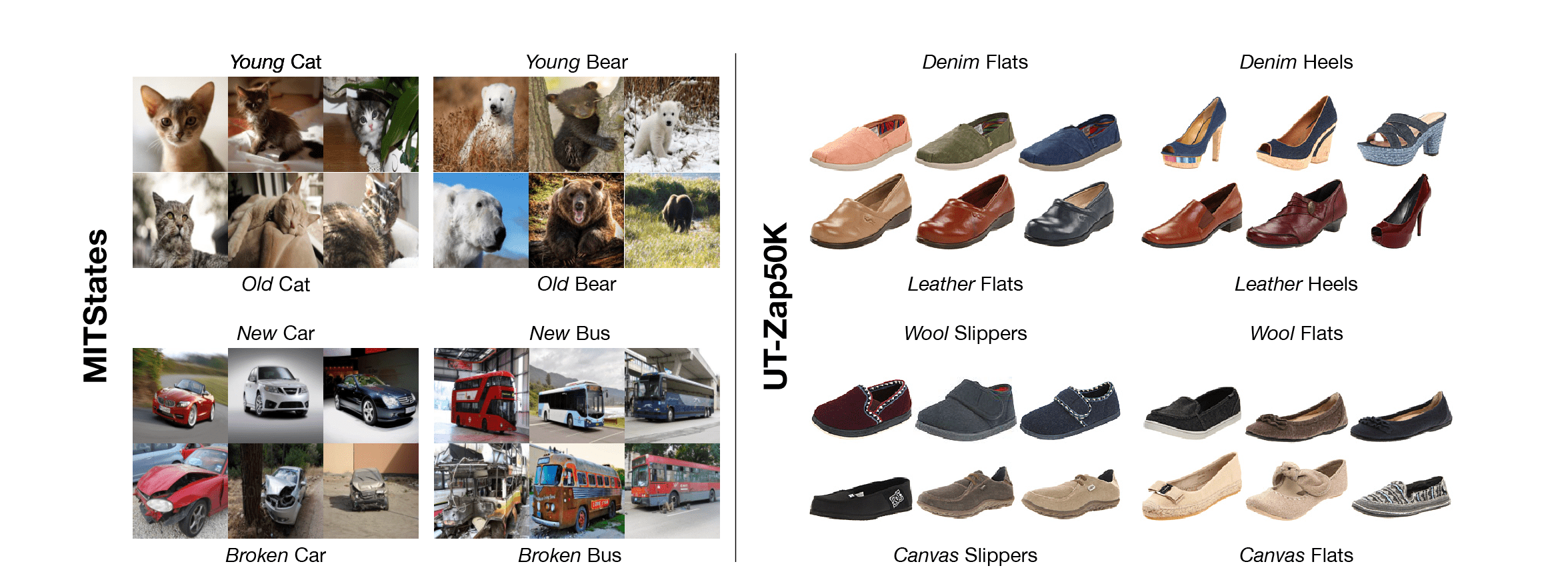}
\caption{Data samples of the MIT-States and UT-Zap50K datasets. An attribute-object composition is associated to each images. Only a subset of the composition is seen during training. Both MIT-States and UT-Zap50K are fine-grained recognition datasets where images in MIT-States come from natural scenes while images in UT-Zap50K are mostly with white background, depicting shoes with different materials.}
\label{fig:data_sample}
\end{figure*}
\minisection{Datasets.} 
We conduct experiments on three datasets: MIT-States~\cite{isola2015discovering}, UT-Zap50k~\cite{yu2017semantic} and StanfordVRD~\cite{lu2016visual}. For
MIT-States, samples of which are shown in Figure~\ref{fig:data_sample} left, each image is associated with an attribute-object pair, e.g., \textit{modern city}, \textit{sunny valley}, as the label. The model is trained on 34K images with 1,292 labeled seen pairs and tested on 34K images with
700 unseen pairs. The UT-Zap50k dataset (samples shown in Figure~\ref{fig:data_sample} right) is a
fine-grained dataset where each image is associated with a material attribute and shoe type pair
(e.g., \texttt{leather slippers},
\texttt{cotton sandals}). Following~\cite{nagarajan2018attributes}, 25k images of 83 pairs
are used for training and 4k images of 33 pairs for testing. 
We also consider compositions
that go beyond attribute-object pairs. For StanfordVRD, the visual concept 
is represented with a SPO (subject, predicate, object) triplet, e.g., \texttt{person wears jeans}, \texttt{elephant on grass}. The dataset has 7,701 SPO triplets, of which 1,029 are seen only in the test set. Similarly to~\cite{misra2017red}, we crop the images with the 
ground-truth bounding boxes and treat the problem as classification of SPO tuples rather than detection. We obtained 37k bounding box images for training and 1k for testing. 

\minisection{Experimental details.} In the experiments, we extract the image features with ResNet-18 and ResNet-101~\cite{he2016deep} pretrained on ImageNet following~\cite{misra2017red,nagarajan2018attributes,wang2019tafe} and also include the more recent DLA-34 and DLA-102~\cite{yu2018deep} for benchmarking. We report the top-1 accuracy of the unseen compositions following~\cite{misra2017red,wang2019tafe}. 
We use Glove~\cite{pennington2014glove} to convert the attributes and objects into 300-dimensional word embeddings. In practice, the raw word embeddings of attributes and objects are transformed by two 2-layer FC networks $\phi_a$ and $\phi_o$ with the hidden unit size of 1024. $\phi(t)$ is the concatenation of $\phi_a(a)$ and $\phi_o(o)$ used as input to both $G$ and $D$. 

The discriminator $D$ is a 3-layer FC networks with hidden unit size of 1024. For the generator $G$, we use a 4-layer FC network where the hidden unit size of the first three layers is 2048 and the size of the last layer matches the dimension of the target feature dimension. $E_t$ is a single layer FC network with no bias and hidden unit size matching the corresponding feature layer size of the generator. $E_n$ is a 2-layer FC network where the hidden unit size is 1024 in the first layer, matching the corresponding feature layer size of the generator in the second layer. The classifier $f$ is a simple soft-max classifier with one FC layer. We adopt the Adam~\cite{kingma2014adam} optimizer with an initial learning rate of $10^{-5}$ for the embedding network $\phi$ and $10^{-4}$ for the other parameters. We divide the learning rate by 10 at epoch 30 and train the network for 40 epochs in total, reporting the accuracy of the last epoch. The batch size is 128.

\begin{figure*}[t]
    \centering
    \includegraphics[width=\linewidth]{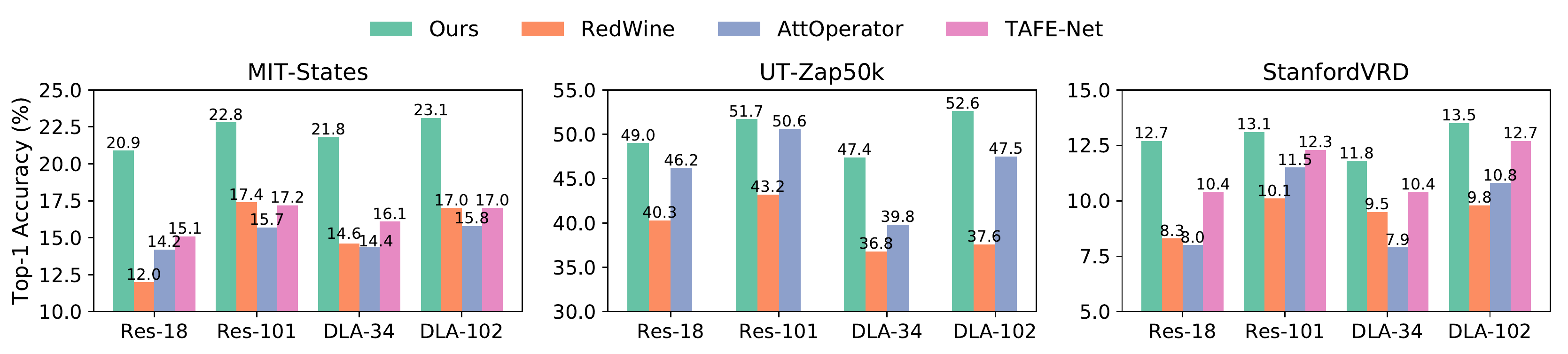}
    \caption{Top-1 accuracy of unseen compositions in compositional zero-shot learning on  MIT-States (700 unseen pairs), UT-Zap50K (33 unseen pairs) and StanfordVRD (1029 unseen triplets). \model (the first bar in each group) achieves state-of-the-art results on all three datasets with four different feature extractors (ResNet-18, ResNet-101, DLA-34 and DLA-102).}
    \label{fig:benchmark_res}
\end{figure*}

\minisection{Quantitative results.} We present the top-1 accuracy of the unseen attribute-object pairs in 
Figure~\ref{fig:benchmark_res} following~\cite{misra2017red,nagarajan2018attributes,Nan2019RecognizingUA,wang2019tafe}. 
We consider three top performing models as our baselines. 
Redwine~\cite{misra2017red} leverages the 
compatibility of extracted generic image features $\mathcal{Z}$ and the task descriptions $\mathcal{T}$ with
a simple binary cross entropy (BCE) loss. 
AttOperator~\cite{nagarajan2018attributes} proposes to use attributes features to modify the object features building on top of the extracted features. It also adopts a metric-learning approach to score the compatibility of transformed image feature as well
as the task embeddings. In our experiments, we report the results of these two methods on different backbone feature extractor using the open-sourced code from Nagarajan and Grauman~\cite{nagarajan2018attributes}\footnote{\url{https://github.com/Tushar-N/attributes-as-operators}}.
TAFE-Net~\cite{wang2019tafe} is a recent method that learns a task-aware feature embeddings for a shared binary classifier to classify the compatibility of task-aware image feature embeddings and the task embeddings. We obtain the benchmark results using DLA as the feature extractor through the released official code\footnote{\url{https://github.com/ucbdrive/tafe-net}}. Our classifier is directly trained on the synthetic image features of the unseen compositions and at testing time, only the real image features of the unseen compositions are fed into the classifier, not combined with the task descriptions as the existing approaches do. 

We present the qualitative results in Figure~\ref{fig:benchmark_res}.  As we can observe 
from the bar charts, our model (denoted as the green bar, the first bar in each group) outperforms the other baseline methods by a large margin on both MIT-States and UT-Zap50k.
Extending from attribute-object pairs to (subject, predicate, object) triplets, our model
also outperforms all the considered baselines. This indicates that \model effectively synthesizes the real image feature distributions of the novel compositions and helps the
classifier generalize to novel concepts without using real image features.

\subsection{Generalized Zero-Shot Compositional Learning}
\label{sec:gcz}
In this section, we provide evaluation on the \emph{generalized} zero-shot compositional learning recently introduced by Purushawakam \textit{et al.}~\cite{purushwalkam2019task}. As pointed by Purushawakam \textit{et al.}, the previous zero-shot compositional learning benchmark does not carefully evaluate the overall system performance when balancing both the seen and unseen compositions. Therefore, they introduce new data splits of the MIT-States and UT-Zap50k datasets and adopt the AUC value as the evaluation metric to examine the calibrated model performance. Our model is able to outperform the previous methods by a large margin with an over 2$\times$ accuracy on the MIT-States dataset.

\minisection{Data splits.} In this generalized ZSCL task, the two datasets (MIT-States and UT-Zap50k) have the same images as used in the ZSCL task. In the new data split, 
the training set of MIT-States has about 30K images of 1262 compositions (the
\emph{seen} set), the validation set has about 10K images from 300 seen and 300 unseen compositions. The testing set has about 13K images from 400 seen and 400 unseen compositions.  On the UT-Zap50K dataset, which has 12 object classes and 15 attribute classes, with a total of 33K images. The dataset is split into a training set containing about 23K images of 83 seen compositions. The validation set has about 3K images from 15 seen and 15 unseen compositions. The testing set has about 3K images from 18 seen and 18 unseen pairs. 

\minisection{Metric.} Instead of using the top-1 accuracy of the unseen compositions, 
Purushawakam \textit{et al.}~\cite{purushwalkam2019task} introduce a set of calibration biases (single scales added to the scores of all unseen pairs) to calibrate the implicit bias imposed to the seen compositions during training. For a given value of the calibration bias, accuracies of both the seen and unseen compositions are computed. Because the values of the calibration bias have a large variation, we draw a curve of the accuracies of seen/unseen compositions and the area blow the curve (AUC) can describe the overall performance of the system more reliably. 

\minisection{Quantitative results.} Table~\ref{tab:auc} provides comparisons between our model and the previous methods on both the validation and testing sets. The network structures of our model is the same as those used in the ZSCL task
and the best training epochs are decided by the validation set. As Table~\ref{tab:auc} shows,  our model outperforms the previous methods by a large margin.  On the challenging MIT-States dataset which has about 2000 attribute-object 
pairs and is inherently ambiguous, all the baseline methods have a relatively low AUC score while our model is able to double the performance of the previous methods, indicating the effectiveness of our model.

  \begin{table}[t]
    \centering
    
    \caption{AUC in percentage on MIT-States and UT-Zap50K. Our model outperforms the previous methods by a large margin, doubling the performance of the prior art on MIT-States. \label{tab:auc}}
    \setlength\tabcolsep{3pt} 
    \adjustbox{max width=\textwidth}{
    \begin{tabular}{lcccccccccccc}
    \toprule
    & \multicolumn{6}{c}{\textbf{MIT-States}} & \multicolumn{6}{c}{\textbf{UT-Zap50K}} \\
    & \multicolumn{3}{c}{Val AUC} & \multicolumn{3}{c}{Test AUC} & \multicolumn{3}{c}{Val AUC} & \multicolumn{3}{c}{Test AUC} \\
    Model   Top \emph{k}$\rightarrow$ & 1 & 2 & 3 & 1 & 2 & 3 & 1 & 2 & 3 & 1 & 2 & 3 \\
    \midrule
    AttOperator~\cite{nagarajan2018attributes} & 2.5 & 6.2 & 10.1 & 1.6 & 4.7 & 7.6 & 21.5 & 44.2 & 61.6 & 25.9 & 51.3 & 67.6 \\ 
    RedWine~\cite{misra2017red} & 2.9 & 7.3 & 11.8 & 2.4 & 5.7 & 9.3 & 30.4 & 52.2 & 63.5 & 27.1 & 54.6 & 68.8 \\
    LabelEmbed+~\cite{nagarajan2018attributes} & 3.0 & 7.6 & 12.2 & 2.0 & 5.6 & 9.4 & 26.4 & 49.0 & 66.1 & 25.7 & 52.1 & 67.8 \\
    TMN~\cite{purushwalkam2019task} & 3.5 & 8.1 & 12.4 & 2.9 & 7.1 & 11.5 & 36.8 & 57.1 & 69.2 & 29.3 & 55.3 & 69.8 \\ \midrule
    \model (Ours) & \textbf{8.9} & \textbf{18.0} & \textbf{25.5} & \textbf{6.5}  & \textbf{14.0}  & \textbf{20.0} & \textbf{41.1} & \textbf{65.3} & \textbf{78.1} &\textbf{32.4} &\textbf{58.1} & \textbf{70.9}\\
    \bottomrule
     \end{tabular}
      }
  \end{table}

\subsection{Ablation study}
\label{sec:ablation}
In this section, we provide various ablation study on the generator architecture comparing various sampling strategies. We also compare our task-aware deep sampling strategy with the shallow sampling scheme used in CLSWGAN~\cite{xian2018feature}, indicating that our generator design can not only lead to more accurate classification predictions but more sample efficient. In addition, we conduct ablation study over the loss objective used in our model. 
\begin{figure*}[h]
    \centering
    \includegraphics[width=\linewidth]{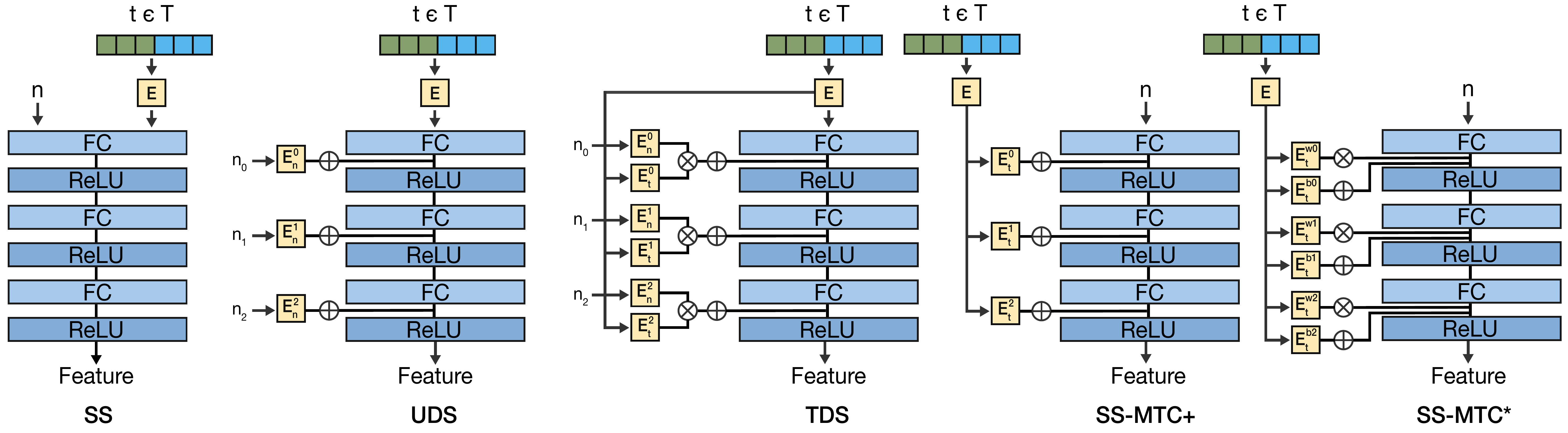}
    \caption{Depictions of various architectures, SS, UDS, TDS, SS-MTC+, SS-MTC*. SS, shallow sampling, does not inject noise at each layer while the rest do. The right three, the task-aware deep sampling (TDS), SS-MTC+, and SS-MTC* inject the task embedding at each layer but only TDS injects task and noise at each layer. We find that our chosen generator design using TDS in the middle of the figure obtains the highest classification accuracy compared to other designs.}
    \label{fig:ablation_arch}
\end{figure*}

\minisection{Alternative generator designs.} We analyze two central
differences of our task-aware deep sampling (TDS) strategy: deep sampling and multi-step
task conditioning. We considered four generator designs using other sampling strategies
as depicted in Figure~\ref{fig:ablation_arch}. The leftmost, shallow sampling (SS) takes noise and the task as input once at the beginning whereas our \model repeatedly injects task-conditioned noise at each layer. Unconditional Deep Sampling (UDS) in contrast injects noise at each layer but does not use the task information at each layer. Two other variants, shallow sampling with multi-step task conditioning (SS-MTC) include SS-MTC$+$, which adds task information at every layer to 
the generator, and SS-MTC$*$, which adopts an affine transformation of the features conditioned on the task at each level inspired by FiLM~\cite{perez2018film} and TAFE-Net~\cite{wang2019tafe}. 

In Table~\ref{tab:ablation}, we present the
top-1 accuracy of the unseen compositions on the three datasets using ResNet-18 as the feature extractor. We observe that both SS-MTC$+$ and SS-MTC$*$ have better performance than the vanilla shallow sampling (SS) with single step task conditioning and that SS-MTC$*$ has better performance than SS-MTC$+$ due to the more complex transformation. In addition, we find the unconditioned deep sampling (UDS) is better than SS, though both of them use single step task conditioning. In all cases, the proposed TDS, which utilizes both deep sampling and multi-step conditioning,
achieves the best results among all the considered variants. 

  \begin{table}[t]
    \centering
    \caption{Top-1 Accuracy of unseen compositions. SS-MTC$+$ and SS-MTC$*$ utilizing multi-step conditioning have better performance than SS. UDS with deep sampling achieves higher accuracies than SS. Overall, Task-aware deep sampling (TDS) achieves better performance than all the alternatives.\label{tab:ablation}}
    \adjustbox{max width=\textwidth}{
    \begin{tabular}{lccc}
    \toprule
    Sampling  & MIT-States & UT-Zap50K & StanfordVRD\\
   Strategy & Top-1 Acc. (\%)  & Top-1 Acc. (\%)  & Top-1 Acc. (\%) \\
    \midrule
    SS & 12.4   & 40.0  &  8.3    \\
    UDS & 14.8 & 41.4 & 8.7 \\
    SS-MTC$+$ & 18.3 & 43.4 &  9.3  \\
    SS-MTC$*$ & 19.2 & 44.3 & 10.1  \\
    \midrule
    TDS & \textbf{20.9} & \textbf{49.0} & \textbf{12.7} \\
    \bottomrule 
     \end{tabular}
      }
  \end{table}

  \begin{figure}[h]
    \centering
    \includegraphics[width=.9\linewidth]{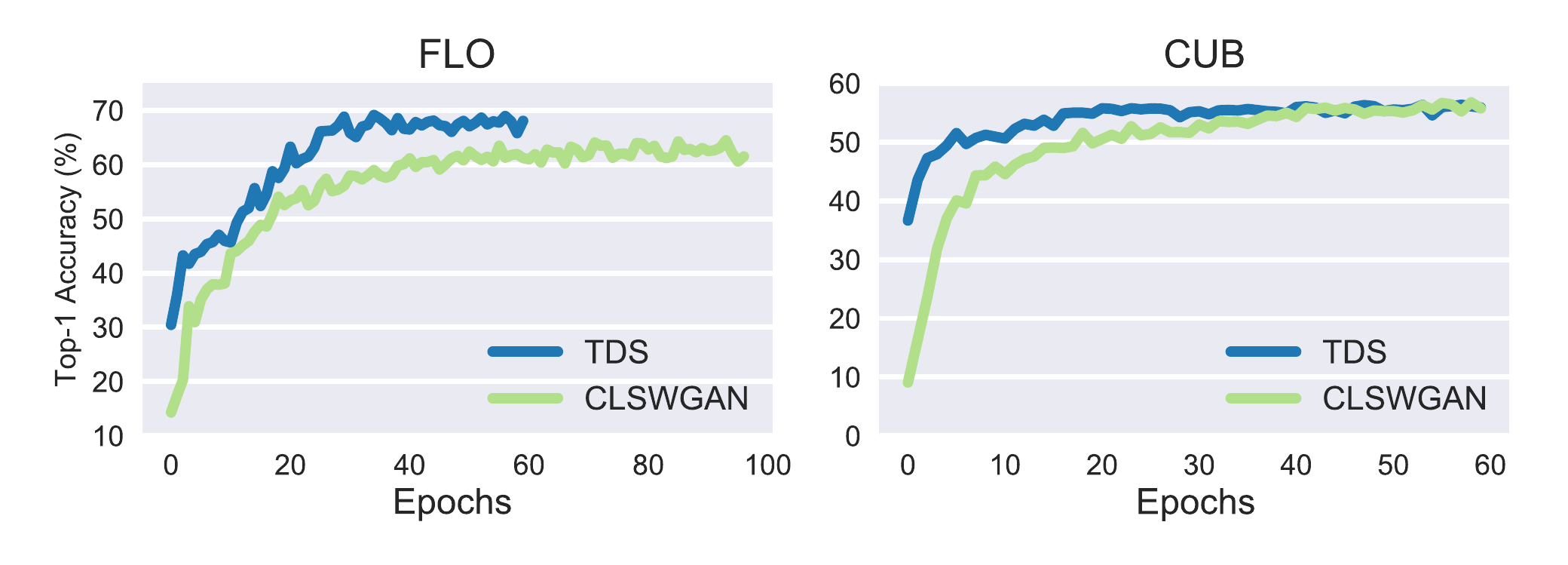}
    \caption{Top-1 test accuracy \emph{vs.} training epochs. \model achieves better sampling efficiency on both FLO and CUB. In particular, \model converges at $\sim$15 epochs while CLSWGAN converges at $\sim$40 epochs on CUB.\vspace{-3mm}}
    \label{fig:conv_zsl}
\end{figure}
\minisection{Faster convergence.} As mentioned in the previous sections, CLSWGAN is one of the closest feature generation work in the zero-shot learning literature. In order to evaluate the generality of the our generator design, we plug in our generator design into the released code of CLSWGAN and compare the model performance in the same datasets (FLO and CUB) and zero-shot learning setting used by CLSWGAN. We observe that our model converges faster than CLSWGAN and has better sample efficiency. 

In Figure~\ref{fig:conv_zsl}, we plot the test accuracy of FLO and CUB under the ZSL setting at every training epoch, the original experimental setting adopted by Xian\textit{ et al.}~\cite{xian2018feature}. Our approach converges at around 30 epochs on FLO and 15 epochs on CUB while CLSWGAN converges at around 40 epochs on both FLO and CUB. We conjecture this is because deep sampling allows for local data sampling at each layer of the generator along a fixed mapping from task descriptions to target image features rather than having to learn a global transformation from the initial random distribution to the target data distribution. Thus it is easier to optimize.

\begin{table}[h]
    \centering
    \caption{Loss ablation. The prediction accuracy drops significantly when removing $\mathcal{L}_\text{cls}$.}
    \adjustbox{max width=\linewidth}{
    \begin{tabular}{l|ccc}
    \toprule
   Model    & MIT-States & UT-Zap50K & StanfordVRD\\
    \midrule
    \model w/o $\mathcal{L}_\text{cluster}$  & 0.33 & 20.2 & 0.17 \\
    \model w/ $\mathcal{L}_\text{cluster}$ & \textbf{20.9} & \textbf{49.0} & \textbf{12.7} \\
    \bottomrule 
    \end{tabular}
    }
    \label{tab:loss}
\end{table}

\minisection{Loss ablation.} In Section~\ref{sec:objective}, we describe our objective function. Besides the commonly used classication loss, adversarial training loss, we additionallyy introduce the clustering loss $\mathcal{L}_\text{cluster}$ as part of the objective function. As shown in Table~\ref{tab:loss}, the accuracy drops significantly if the $\mathcal{L}_\text{cluster}$ is removed. We conjecture
that this regularization helps the synthetic features to capture the statistics, e.g., centroid, of the target data distribution quickly so useful features are generated efficiently to help train the classifier.

\begin{figure*}[h]
    \centering
    \includegraphics[width=.9\linewidth]{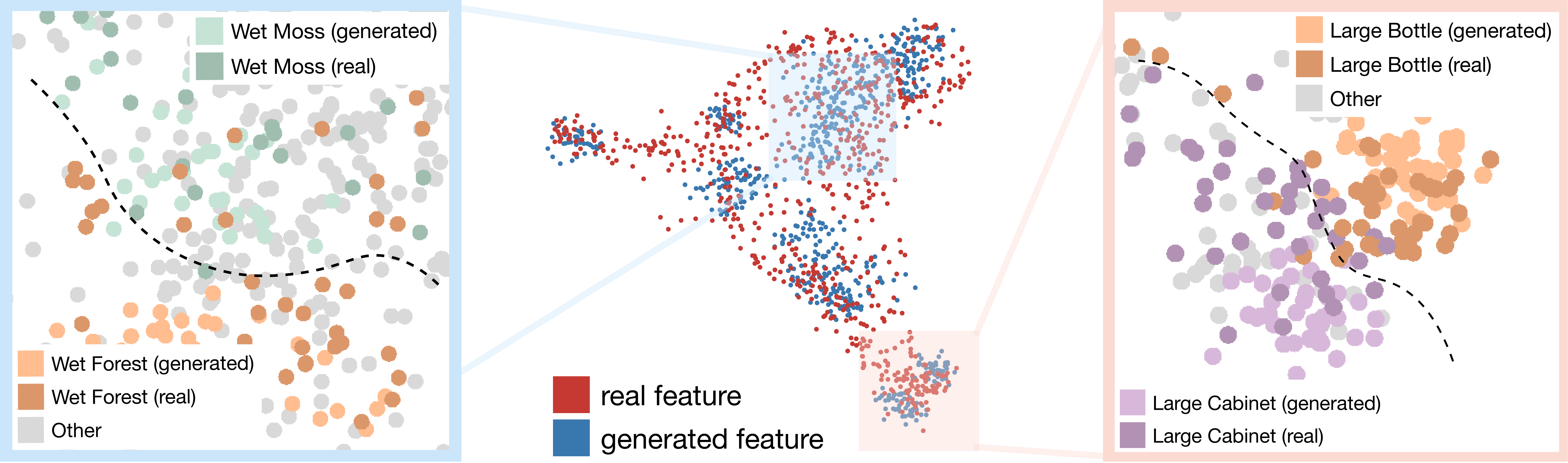}
    \caption{Feature visualization of real and generated features of images in the testing set. The center depicts real features, represented as red points, and generated features, represented by generated features visualized using UMAP. Within different regions, we observe in the left and right, that the generated feature distribution closely matches the real feature distribution, and that distributions of different classes are separated.}
    \label{fig:emb_vis}
\end{figure*} 

\subsection{Qualitative Results}
\label{sec:vis}

\minisection{Visualization of synthetic features.} Observing Figure~\ref{fig:emb_vis} which depicts the real and generated features of the novel compositions on MIT-States, 
we can see that the synthetic features (in blue) overlap with the real features (in red). The synthetic features form rough clusters compared to the real features, 
which may make training of the classifier easier. Zooming in to check different
regions of the feature distributions (in the windows on both sides of Figure~\ref{fig:emb_vis}), we find that though
semantically closer compositions are also closer in the image feature space, e.g., \textit{wet moss} and \textit{wet forest} in the window on the left, the synthetic 
features still closely cover the real feature distribution and form cleaner 
cluster boundaries between different compositions than the real features. 

\begin{figure}[h]
    \centering
    \includegraphics[width=.9\linewidth]{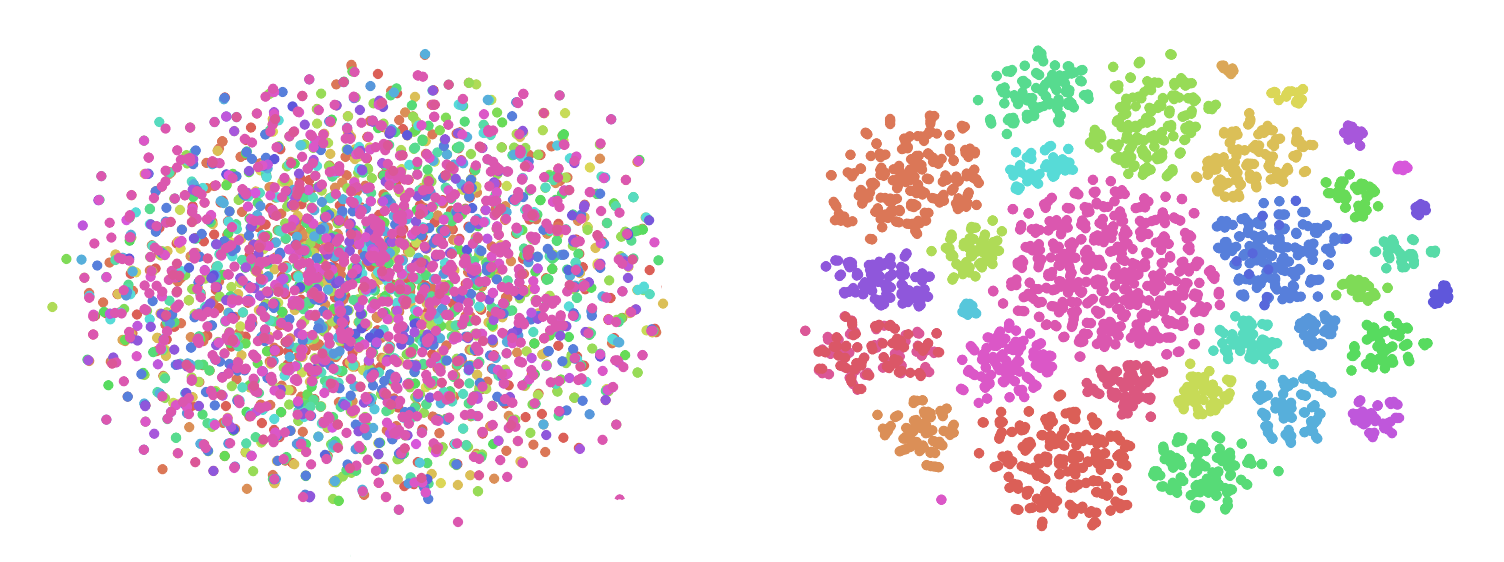}
    \caption{T-SNE visualization of the unconditioned noise used in UDS (left) and task-aware noise injected in the last layer of \model (right) of 33 unseen attribute-object compositions on UT-Zap50K. The task-aware noise is clustered based on the task while the unconditioned noise is mixed in one cluster.\vspace{-2mm}}
    \label{fig:tsne}
\end{figure}

\minisection{Visualization of task-conditioned noise.} As discussed in the previous section, the task-aware deep sampling (TDS) used in our generator design is one of the key components that allow the model to achieve better sampling efficiency. 
TDS is different from the unconditional deep sampling (UDS) mainly because of the injection of
the task-conditioned noise, which allows for sampling from a task-adaptive distribution. 

In Figure~\ref{fig:tsne}, we visualize
the noise injected to the last layer of the generator in UDS and TDS with t-SNE~\cite{maaten2008visualizing} of the 33 unseen compositions on UT-Zappos. We can observe from the figure that the task-aware noise is clustered based on the task while the unconditioned noise is mixed in one cluster. We hypothesize that the task-relevant samples
injected to the generator help the generator to estimate the target image feature
distribution.

\begin{figure*}[t]
\subfigure[MIT-States]{
\centering
    \begin{minipage}{.42\linewidth}
        \includegraphics[height=8\baselineskip]{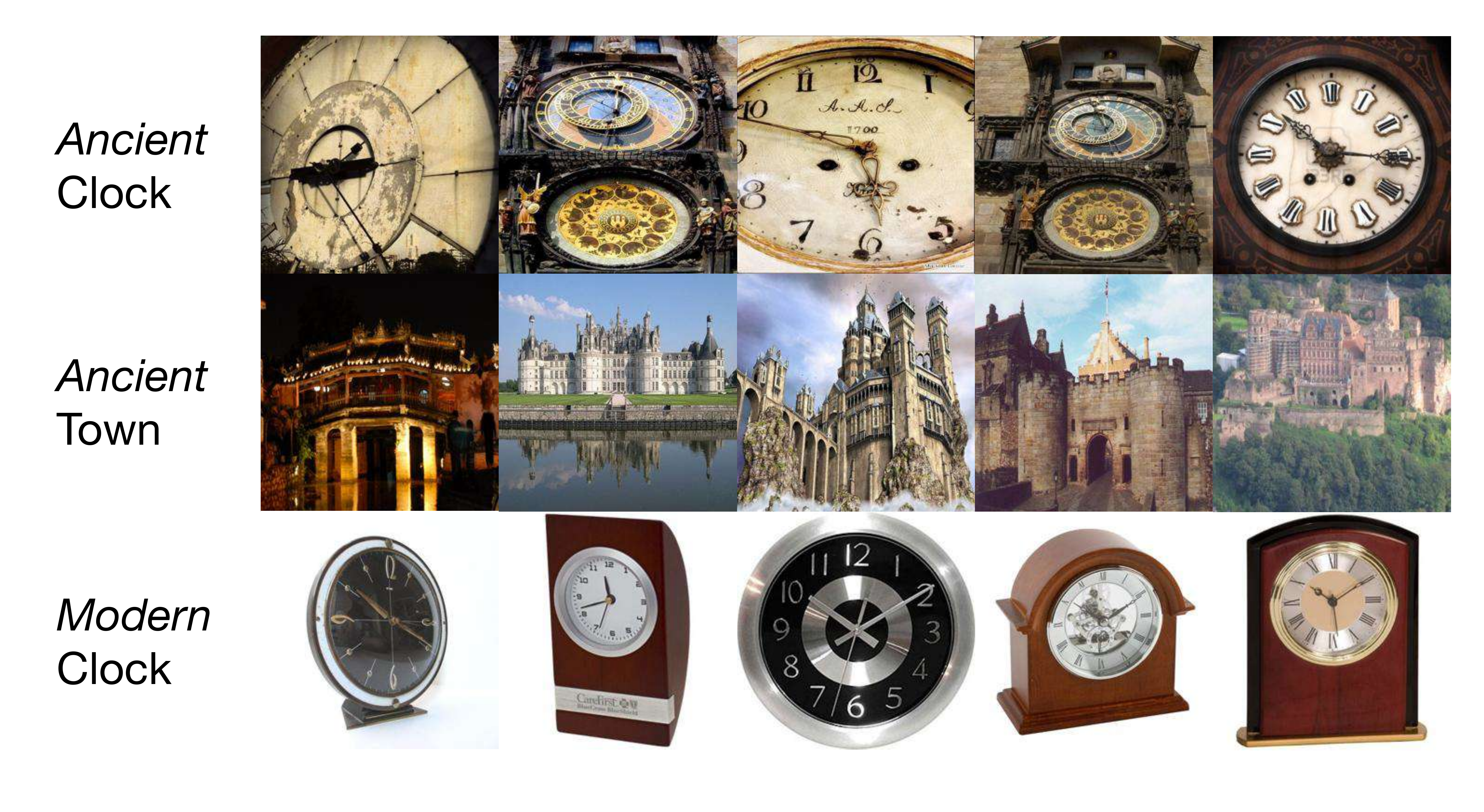}
    \end{minipage}
}
~
    \subfigure[StanfordVRD]{
    
    \begin{minipage}{.42\linewidth}
        \includegraphics[height=8\baselineskip]{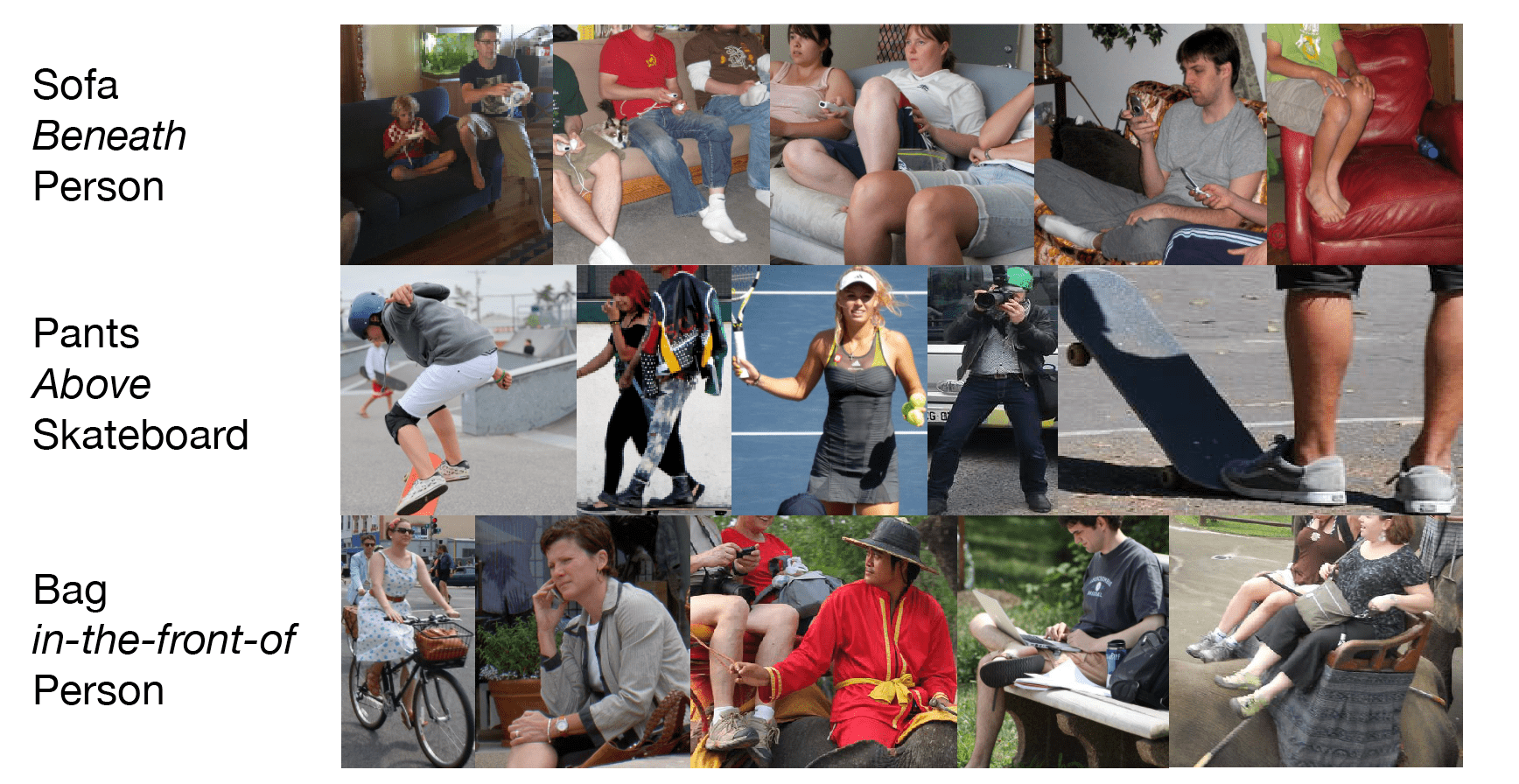}
    \end{minipage}
}
    \caption{Top retrievals of the MIT-States and the StanfordVRD dataset. For different classes, we list the top-5 images that have the highest probability score assigned to those classes. We can see that most of the retrieved images match the corresponding attribute-object compositions. However, failure cases do exist, especially in the challenging StanfordVRD datasets. For example,  there are several mismatches in the second row of the figure on the right. }
    \label{fig:topret}
\end{figure*}

\minisection{Top retrievals of novel compositions.} In Figure~\ref{fig:topret}, we visualize the top retrievals of the novel compositions from the MIT-Statese and the StanfordVRD datasets. 
We pick up the images that the classifier assigns the highest probability score for three different novel compositions in the evaluation set. Based off these images, the classifier appears to match the images to correct novel compositions. However, in some compositions, e.g., Pants Above Skateboard, we observe some mismatching examples. 

\section{Conclusion}
In this paper, we tackled the zero-shot compositional learning task with a compositional feature synthesis approach. We proposed a task-aware feature generation framework, improving model generalization from the generative perspective. We designed a task-aware deep sampling strategy to construct the feature generator, which produces synthetic features to train classifiers for novel concepts in a zero-shot manner. The proposed \model achieved state-of-the-art results on three benchmark datasets (MIT-States, UT-Zap50K and StanfordVRD) of the zero-shot compositional learning task. In the generalized ZSCL task recently introduced by Purushawakam \textit{et al.}~\cite{purushwalkam2019task}, our model is able to improve the previous baselines by over 2$\times$, establishing a new state of the art. An extensive ablation study indicated that the proposed \model can not only improve the prediction accuracy but lead to better sample efficiency and faster
convergence. A visualization of the feature distributions showed 
the generated features closely model the real image feature distributions with clearer separation between different compositions. 

\subsubsection*{Acknowledgments}
This work was supported by Berkeley AI Research, RISE Lab, Berkeley DeepDrive and DARPA. We also thank Andy Yan for the helpful discussion and suggestions on the paper writing.

\bibliographystyle{splncs04}
\bibliography{references}

\end{document}